\documentclass[letterpaper, 10 pt, conference]{ieeeconf}  

\IEEEoverridecommandlockouts                              

\usepackage{graphics} 
\usepackage[demo]{graphicx}
\usepackage{subcaption}
\usepackage[normalem]{ulem} 
\usepackage{amsmath} 
\usepackage{amssymb}
\usepackage{mathcomp}
\usepackage{array}
\newcolumntype{M}[1]{>{\centering\arraybackslash}m{#1}}
\usepackage{booktabs}
\usepackage{multirow}
\usepackage{siunitx}
\usepackage{caption}
\usepackage{ulem}
\usepackage{color}
\usepackage{kotex}
\usepackage{ulem}
\usepackage{verbatim}
\usepackage{tabularx}
\newcolumntype{L}[1]{>{\raggedright\arraybackslash}p{#1}}
\newcolumntype{C}[1]{>{\centering\arraybackslash}p{#1}}
\newcolumntype{R}[1]{>{\raggedleft\arraybackslash}p{#1}}
\usepackage{hyperref}
\usepackage{cleveref}
\hypersetup{
    colorlinks=true,
    linkcolor=black,
    filecolor=magenta,      
    urlcolor=cyan,
}
 
\makeatletter
\newcommand{\thickhline}{%
    \noalign {\ifnum 0=`}\fi \hrule height 0.8pt
    \futurelet \reserved@a \@xhline
}
\newcolumntype{"}{@{\hskip\tabcolsep\vrule width 0.8pt\hskip\tabcolsep}}
\makeatother

\title{\LARGE \bf
Moving object detection for visual odometry in a dynamic environment based on occlusion accumulation 
}

\author{Haram Kim$^1$ , Pyojin Kim$^2$ and H. Jin Kim$^1$
\thanks{The SNU-Samsung smart campus research center (0115-20190023) at Seoul National University provides research facilities for this study.}
\thanks{$^1$ Lab for Autonomous Robotics Research (LARR), Seoul National University, Seoul, South Korea} 
\thanks{$^2$ Computer Graphics and Vision Lab (GrUVi), Simon Fraser University, Burnaby, BC}
}

\begin{document}

\maketitle
\thispagestyle{empty}
\pagestyle{empty}

\begin{abstract}

 Detection of moving objects is an essential capability in dealing with dynamic environments. Most moving object detection algorithms have been designed for color images without depth.
For robotic navigation where real-time RGB-D data is often readily available, utilization of the depth information would be beneficial for obstacle recognition.
 Here, we propose a simple moving object detection algorithm that uses RGB-D images. The proposed algorithm does not require estimating a background model.
Instead, it uses an occlusion model which enables us to estimate the camera pose on a background confused with moving objects that dominate the scene.
The proposed algorithm allows to separate the moving object detection and visual odometry (VO) so that an arbitrary robust VO method can be employed in a dynamic situation with a combination of moving object detection, whereas other VO algorithms for a dynamic environment are inseparable. In this paper, we use dense visual odometry (DVO) as a VO method with a bi-square regression weight. Experimental results show the segmentation accuracy and the performance improvement of DVO in the situations. We validate our algorithm in public datasets and our dataset which also publicly accessible.

\end{abstract}

\section{Introduction}
Dynamic environments still pose a challenge in robotics. A capability to detect moving objects allows extending results of various studies performed for static scenes to dynamic environments. Although much research such as the background subtraction method has been carried out to recognize moving objects, it is ineffective in scenes where moving objects occupying a major portion of the image,  {(which we refer to as ``moving objects \textit{dominate} the scene").} 
Such situations happen frequently when moving objects are nearby and should be treated as a top priority. 

\begin{figure}[t]
 \centering
\begin{subfigure}{.22\textwidth}
  \centering
  \includegraphics[width=.9\textwidth]{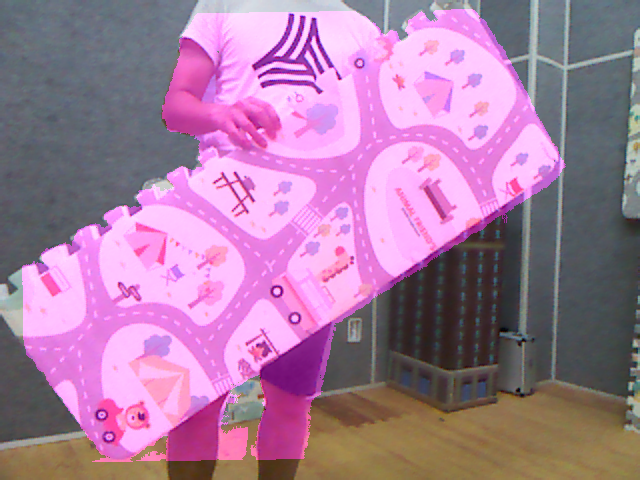}
  \caption{Object detection result}
  \label{fig:first_sfig3}
\end{subfigure} %
\begin{subfigure}{.22\textwidth}
  \centering
  \includegraphics[width=.9\textwidth]{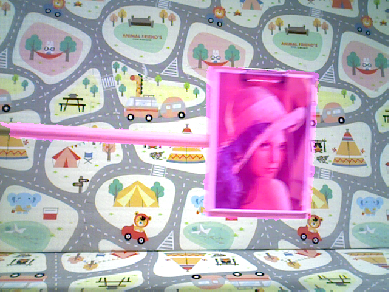}
  \caption{Object detection result}
  \label{fig:first_sfig4}
\end{subfigure} \\
\vspace{0.15cm}
\begin{subfigure}{.22\textwidth}
  \centering
  \includegraphics[width=.9\textwidth]{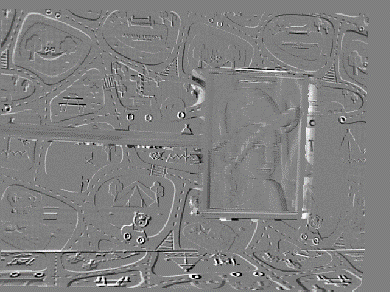}
  \caption{Residual image from the robust DVO}
  \label{fig:first_sfig5}
\end{subfigure} %
\begin{subfigure}{.22\textwidth}
  \centering
  \includegraphics[width=.9\textwidth]{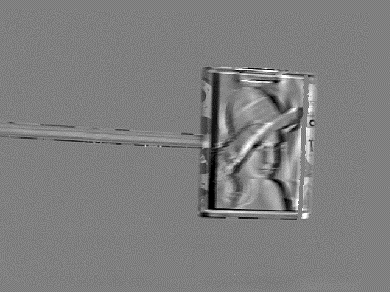}
  \caption{Residual image from the proposed method }
  \label{fig:first_sfig6}
\end{subfigure} 

\caption{ Results of moving object detection. (a), (b) are the results of detecting moving objects, and (a) contains a large moving object which dominates the scene with a freely moving camera. (c) Typical DVO only approach  reaches a local minimum whereas (d) the proposed method demonstrates that the global optimum to the background is reached.}
\label{fig:first}
\vspace{-0.7cm}
\end{figure}
To address the above-mentioned problems arising from dynamic environments, we propose an occlusion accumulation method. The proposed method utilizes both depth and camera pose information obtained from VO to detect moving objects without a particular background model. Some direct VO methods using robust regression weight \cite{Kerl2013ROE} are not able to track the accurate camera pose by falling into a local minimum when a moving object dominates the scene. In this work, we use the estimated camera pose up to the previous image to distinguish the moving objects and exclude them from the images to find the next camera pose. {If only the robust regression in the optimizer (bi-square regression in this paper) is used, the camera pose is likely to be estimated with respect to the moving object when the moving object dominates the scene. In the proposed method that estimates the camera pose with robust regression, the method excludes the moving object before it dominates the scene. Unlike the other mapping methods where memory usage increases each time an image is processed, the memory requirement of the proposed method is not heavy regardless of a long duration of image sequences, because it continuously updates the single occlusion information. } 


We warp the previous depth image using the estimated camera pose and subtract with the current depth image. Accumulating those subtracted values shows a similar result to the background subtraction without relying on background modeling. Our method is robust to the effect of the moving object which dominates the scene and avoids other problems that occur when the background is incorrectly modeled. It is possible to detect a moving object which was originally static and regarded as the background. 

The purpose of our research is to detect moving objects and to improve the performance of arbitrary VO in dynamic environments. Our contributions are summarized as follows.

\begin{itemize}

\item  We suggest {a new method of the occlusion accumulation} which can detect moving objects that dominate the scene, without a background model.  
{\item Our algorithm can improve the performance of any VO algorithm in a dynamic environment by augmenting it.}
\item The proposed algorithm can detect objects which have similar texture to the background and can be further improved by using an externally obtained camera pose.

\end{itemize} 

\section{Related Work}

There are several types of moving object detection algorithms 
\cite{Mehran2018NTMOD} using background modeling, trajectory classification, etc. The methods in \cite{Amitha2015BMBC,EEDMTF,MODTTVCMC,BSMCAG}, belonging to the  category of background modeling, commonly extract feature points and clustered features to distinguish the background and the foreground. They warp the previous image with homography transform attained from features in the dominant cluster.  Such methods show a quite accurate object segmentation result in the scenes dominated by the background, but they assume that the dominant cluster is from a background, and they take a few seconds to process. 

In \cite{Yi2013DMONC}, the background and foreground are distinguished by a dual-mode single Gaussian model (SGM). The adaptive adjustment of the SGM learning rate reduces computation time, but some moving objects are perceived as background if the background is similar in color to the object.

In \cite{BSFMC},  trajectory classification is conducted. They collect points that are tracked for several frames and find the dominant camera motion by applying RANSAC on the trajectories for tracked points. The authors defined conditional probability on trajectory models and applied the maximum a posterior rule to distinguish the points included in the moving object. The authors of \cite{BSMCBTSLI, OSLTAPT} also propose a trajectory classification method by defining a low-dimensional descriptor for describing trajectory shape or spectral clustering with spatial regularity. They detect the outlier trajectories whose sum of the Euclidean distance to trajectories of the nearest neighbors exceeds a specified threshold or by using K-means clustering. The methods of \cite{BSFMC, BSMCBTSLI, OSLTAPT} 
require an appropriate feature extraction method, and they may show sparse segmentation results due to utilization of the tracked features.

Because the aforementioned methods aimed to detect moving objects or tracking objects far from the camera, they used datasets where the background dominates the scene such as a surveillance system or sports relay. On the other hand, in robotic applications, because it is important to process in real-time and detect nearby objects which may occupy the scene, these studies can become less effective.

With the growing popularity of RGB-D cameras such as Kinect and ASUS Xtion PRO, moving object detection algorithms have begun to utilize them. The method in \cite{MDTSA}  estimates camera ego-motion using depth values, and then it uses a homogeneous transformation to warp the previous frame as in \cite{Amitha2015BMBC}, \cite{EEDMTF}, and \cite{MODTTVCMC}. The segmentation is achieved by  the particle filter and vectorization method. Accurate image warping is performed using depth values, but the problem of objects occupying a major portion of the image is not addressed in these methods. In \cite{Javed2017OSRPCA}, they calculated a spatiotemporal graph Laplacians and spatial smoothness among the background pixels. Then, they adapt RPCA to incorporate two approaches for classifying foreground from background. Due to the high computational complexity of RPCA, computation time per frame is several seconds.

In the field of visual odometry, many studies have investigated  camera pose estimation in dynamic environments by reducing the effect of the moving objects. In \cite{Kerl2013ROE, RRVOD} and \cite{kerl2012},  the camera pose is estimated by assigning the t-distribution or  Huber norm weights, so that the dense visual odometry (DVO) method performs properly in some dynamic environments. The method of \cite{Kim2016EBMB} labels the background with a non-parametric model and depth images obtained from the RGB-D camera, and estimates the camera pose by modifying the energy-based DVO. The algorithm would be ineffective when used for object detection, because they focus on labeling the background to estimate pose stably and does not label the moving object.

The method suggested in \cite{leereal} performs rigid motion segmentation on RGB-D camera by a grid-based optical flow. The process of flow calculation and spatial, temporal segmentation was conducted in real-time, but the resulting segmentation  \cite{leereal} was a low resolution that is not as clear as the silhouette-shaped segmentation.

In \cite{Jaimez2017FOSF}, the authors propose a robust camera pose estimation in dynamic environments and 3D scene-flow estimation on the moving object by applying depth geometric clustering and checking whether clusters match the camera motion of images. 
They labeled the foreground and background as probability, and utilize the probability label as a weighing factor on pose estimation to perform well in dynamic environments. The algorithm does not clearly distinguish moving objects and the specified number of clusters would affect the performance. 


\section{Moving Object Segmentation}
 Various factors such as illumination change, moving object, camera ego-motion induce image changes. Assuming that the photo-consistency assumption is not violated, in a static scene that involves moving objects with fixed camera, image change indicates the moving objects. In a dynamic scene which has arbitrary camera motion, we can attain a motionless image by warping the previous image with the estimated camera pose. We use the camera pose estimated from DVO with robust regression weight in this paper. We will refer to DVO with robust regression weight as robust DVO in this paper and more details about robust regression weight will be covered in Section \ref{pose_esti}. After attaining a motionless image, the dynamic scene can be treated as a static scene. We focus on this feature and detailed explanation will follow.
 
\subsection{Occlusion Accumulation} 
 When a moving object roams in the scene, depth can be changed. The depth change can be classified into two kinds. The first one is occlusion: the object will occlude the background in the head of the direction of movement. The second one is reappearance: the background will appear in the tail of the direction of movement. We assume that the depth of background is larger than that of the moving object. Object which pass the rear of the static background such as pillar, tree, etc. do not affect the depth value. We consider the pixel whose depth value changes to a smaller value as a pixel of the moving object.

 The occlusion map and the warping function are defined as
\begin{equation} \label{def_occ}
    \Delta Z_{i}(u,\xi_{i}^{i+1}) = Z_i(\text{w}(u,\xi_{i}^{i+1})) - Z_{i+1}(u)
\end{equation}
and
\begin{equation} \label{def_warp}
    \text{w}(u,\xi_i^{i+1}) = \pi (\text{exp}(\xi_i^{i+1})\pi^{-1}(u,Z_i(u)))
\end{equation}
 where $Z_{i}(u)$ is the \textit{i}th depth image and $u$ is a pixel on image $\Omega (640 \times 480)$. $\xi_i^{i+1} \in \mathbb{R}^6 $ represents the 6-DOF camera pose parameter between the \textit{i}th frame and  the (\textit{i}+1)th frame. The project function $\pi: \mathbb{R}^3 \xrightarrow{}\mathbb{R}^2$ maps a 3D point into a 2D image pixel. The term $ \text{exp}(\xi_i^{i+1}) \in \text{SE}(3) $ represents the transformation matrix for the corresponding camera pose parameter $\xi_i^{i+1}$.

 If $\Delta Z_{i} > 0$, it means that the depth has become smaller, i.e. the occlusion has occurred. Otherwise, $\Delta Z_{i} < 0$ means that the depth has increased, i.e. the reappearance. 
 We define the occlusion accumulation map by
\begin{align} \label{def_accum}
   A_{i+1} (u)  &= \Delta Z_{i}(u,\xi_{i}^{i+1}) + \Tilde{A}_{i} (\text{w}(u,\xi_{i}^{i+1}))\nonumber \\
            &\approx \Delta Z_{i}(u,\xi_i^{i+1}) + \sum_{k=1}^{i} \Delta Z_{k}(\text{w}(u,\xi_{k+1}^{i}),\xi_k^{k+1})\nonumber \\
            &\approx  Z_{k_u}(\text{w}(u,\xi_{k_u}^{i+1})) - Z_{i+1}(u)
\end{align}
 The initial occlusion accumulation map is set as $A_{1}(u)=0$ for $\forall u \in \Omega$ and $k_u$ is the index of the frame which first observed the pixel $u$ of the current frame. The $\Tilde{A}(u)$ is truncated $A(u)$ and described later in Eqs.\eqref{def_accum_re1} and \eqref{def_accum_re2},

 The warped pixels before the index at $k_u$ would be out of the observation window $\Omega$ and there is nothing to calculate. The value of $\Delta Z$ would reveal zero values for these indices.

 \begin{figure}
 \centering
\begin{subfigure}{.22\textwidth}
  \centering
  \includegraphics[width=.9\textwidth]{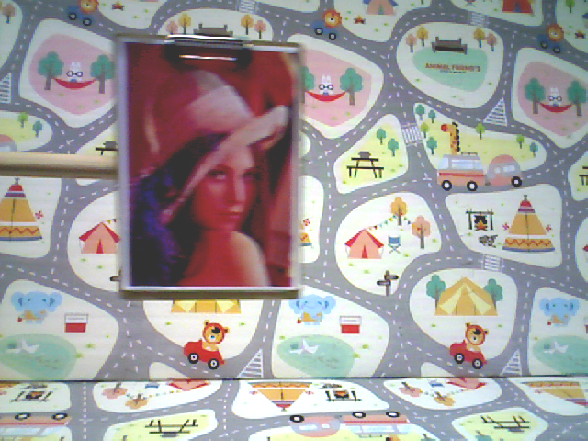}
  \caption{Color image}
  \label{fig:sfig1}
\end{subfigure} %
\begin{subfigure}{.22\textwidth}
  \centering
  \includegraphics[width=.9\textwidth]{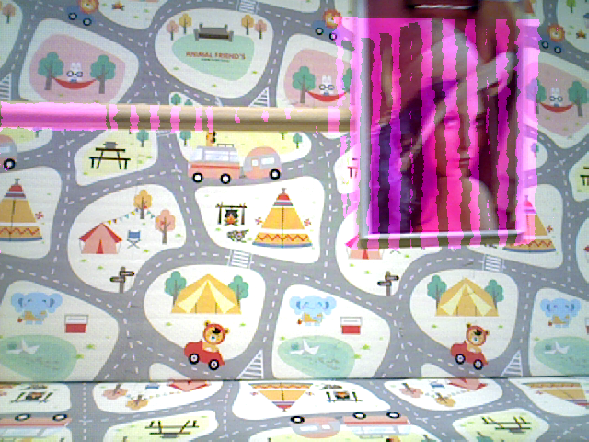}
  \caption{Masked image}
  \label{fig:sfig2}
\end{subfigure} \\
\vspace{0.15cm}
\begin{subfigure}{.22\textwidth}
  \centering
  \includegraphics[width=.9\textwidth]{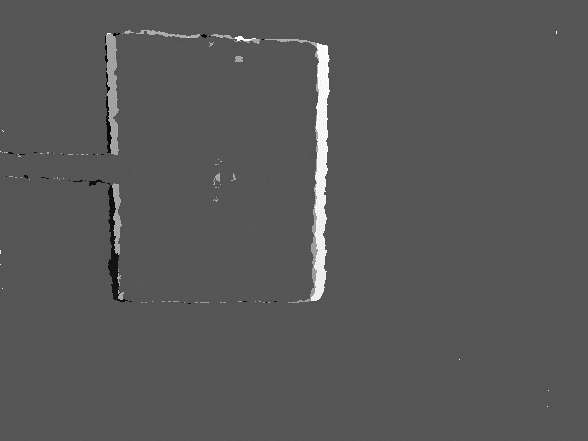}
  \caption{$\Delta Z(u)$}
  \label{fig:sfig3}
\end{subfigure} %
\begin{subfigure}{.22\textwidth}
  \centering
  \includegraphics[width=.9\textwidth]{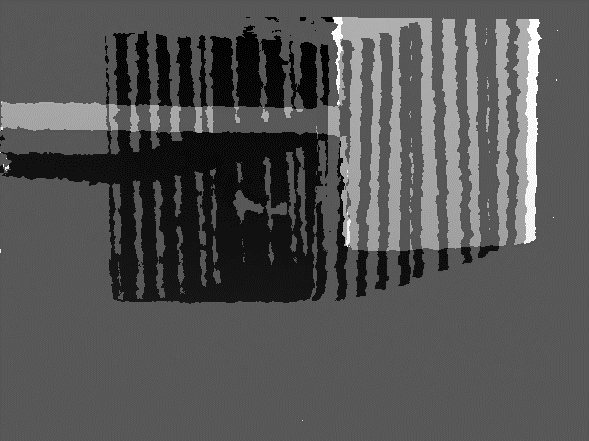}
  \caption{$A(u)$}
  \label{fig:sfig4}
\end{subfigure}
\caption{Illustration of the occlusion accumulation method. $A(u)$ represents the accumulation of $\Delta Z(u)$ over time. A bright area is a region where $A(u)$ shows a positive value and is considered as a moving object, and a dark area is a region where $A(u)$ shows a negative value and the background is revealed again. The middle gray represents a value of zero. The moving object detection results from Eq.\ref{def_bg} is shown in (b). The area considered as a moving object is painted purple.}
\label{fig:accum}
\vspace{-0.5cm}
\end{figure}
 
 When the truncation steps for $A(u)$ have not been triggered, the second and third approximate equalities in Eq.\eqref{def_accum} would be satisfied by $A(u)=\Tilde{A}(u)$. One can interpret the occlusion accumulation map as the result of subtraction between $k_u$th frame and current frame.  In contrast to the background subtraction, we utilize the occlusion map, $\Delta Z_{i}(u,\xi_i^{i+1})$. Thus, our method does not need to depend on recognition of the background or 3D mapping.

 Existing depth sensors have the depth error, which quadratically grows along the depth value in \cite{Jaimez2017FOSF}. To deal with such measurement uncertainty, we set a threshold as 
\begin{equation} \label{def_object}
    \tau_{\alpha}(u) = \alpha \cdot Z(u)^2
\end{equation}
Although we set the threshold to be larger than zero, unwanted error values can exceed the threshold through the accumulation. Furthermore, negative values can be accumulated, so that occlusion could not exceed the threshold at all. In order to reduce their effect, we truncate the occlusion accumulation map $A_i (u)$ to $\Tilde{A}_i (u)$  by the following:
\begin{eqnarray} \label{def_accum_re1}
   \Tilde{A}_i (u)  &=& 0 \qquad \text{for  } A_i (u) \leq \tau_{\alpha}(u)
\\ \label{def_accum_re2}
   \Tilde{A}_i (u)  &=& 0 \qquad \text{for  } \Delta Z_{i}(u,\xi_i^{i+1}) \leq -\tau_{\beta}(u) \;.
\end{eqnarray}
The truncation step in Eq.\eqref{def_accum_re1} could disturb detecting the moving object which slowly approaches toward the camera. However, once the moving object is detected, the small values of $\Delta Z_{i}(u,\xi_i^{i+1})$ will be reflected. When the background reappears, $A_i (u)$ would not be lower than the threshold $\tau_{\alpha}(u)$ due to depth error accumulation. The truncation step in Eq. \eqref{def_accum_re2} helps to detect the background reappearance. The moving object that moves away from the camera hardly exceed the threshold. Even so, if it exceeds the threshold, the object would soon fade away in the scene.
The result of the occlusion accumulation is shown in Fig. \ref{fig:accum}. In order to help understanding, $A(u)$ has not been truncated using Eq. \eqref{def_accum_re1} through every iteration yet. Since the depth value is not measured near the object, a stripe-like result appears in Fig. \ref{fig:sfig4}.

 After calculating $A_i (u)$, we truncate the occlusion accumulation map to distinguish moving objects and background. The background map $B_i (u)$ would reveal 0 if a moving object shows proper occlusion. Otherwise, if the moving object has passed and the background reappears again, $B_i (u)$ would reveal 1, i.e., 
 \begin{equation} \label{def_bg}
    B_i (u) = 
\begin{cases}
    0              & \text{if } A_i (u) > \tau_{\alpha}(u) \\
    1              & \text{otherwise}
\end{cases}
\end{equation}
 The object mask can be calculated by inverting $B_i (u)$.

\subsection{Depth Compensation}
\begin{figure}
 \centering
\begin{subfigure}{.22\textwidth}
  \centering
  \includegraphics[width=.9\textwidth]{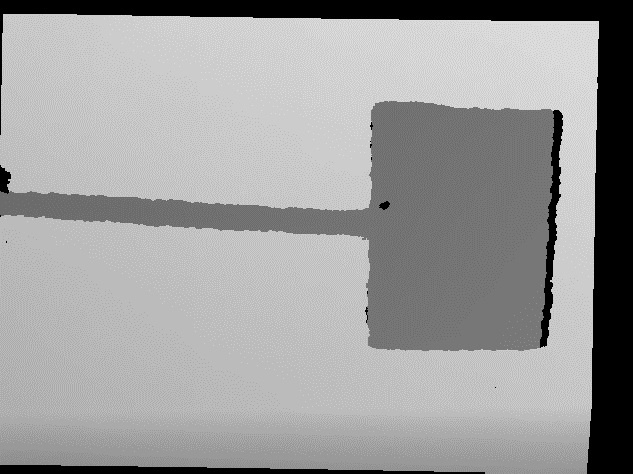}
  \caption{Original depth image}
  \label{fig:com_sfig1}
\end{subfigure} %
\begin{subfigure}{.22\textwidth}
  \centering
  \includegraphics[width=.9\textwidth]{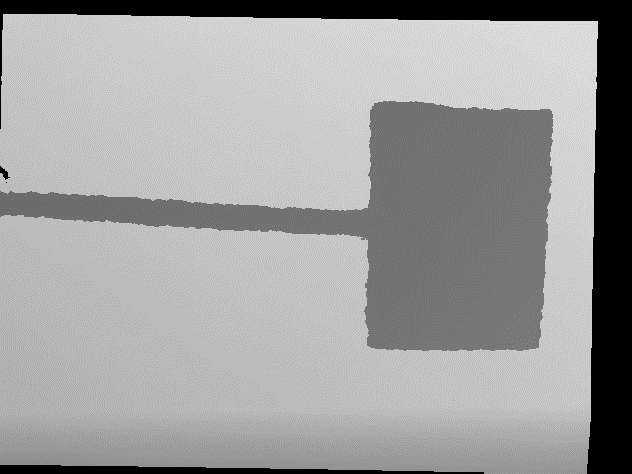}
  \caption{Compensated depth image}
  \label{fig:com_sfig2}
\end{subfigure} \\
\vspace{0.15cm}
\begin{subfigure}{.22\textwidth}
  \centering
  \includegraphics[width=.9\textwidth]{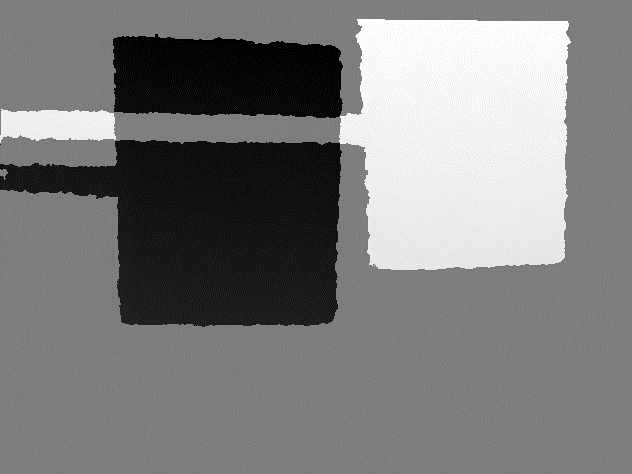}
  \caption{$A(u)$}
  \label{fig:com_sfig3}
\end{subfigure} %
\begin{subfigure}{.22\textwidth}
  \centering
  \includegraphics[width=.9\textwidth]{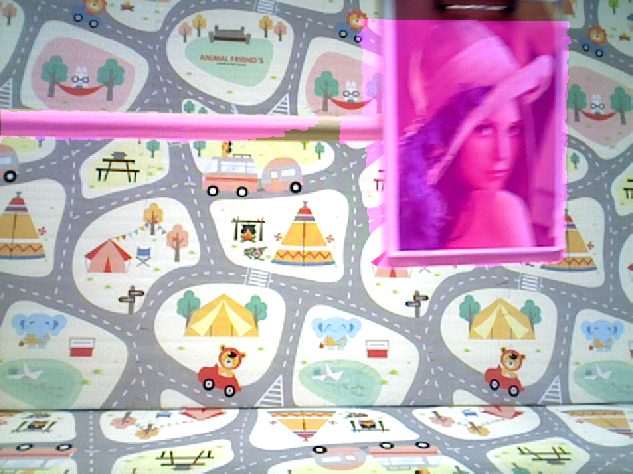}
  \caption{Masked image}
  \label{fig:com_sfig4}
\end{subfigure}
\caption{Result of depth compensation. A raw depth image (a) is compensated as (b). Compared to Fig.\ref{fig:first}, the unwanted stripe-like result is corrected.}
\label{fig:compen}
\vspace{-0.5cm}
\end{figure}

Often, the depth images have invalid depth on an edge of the object or near the border of an image window. Its effect is shown in Fig. \ref{fig:sfig4}. The segmentation in Fig. \ref{fig:sfig2} has empty spaces over the moving object. Some of those invalid depth values in the current frame can be compensated by
\begin{equation} \label{depth_compen}
   Z_i(\Bar{u}) = Z_{i-1}(\text{w}(\Bar{u},\xi_{i}^{i-1}))
\end{equation}
where the unmeasured depth on the pixel $\Bar{u}$  satisfies $Z(\Bar{u})=0$. The compensated depth is used in the next depth image. The result of the depth compensation is shown in Fig.~ \ref{fig:compen}. The original depth image has the  unmeasured area that is colored as black,  near the border and the boundary of the object. The unmeasured area on boundary of the object is filled with the previous depth image as in Fig. \ref{fig:com_sfig2}. From the compensated depth image, $A(u)$ is now calculated reasonably (Fig. \ref{fig:com_sfig3}), so that the object segmentation could cover the entire moving object (Fig. \ref{fig:com_sfig4}).

\subsection{Occlusion Prediction on Newly Discovered Area}
Suppose that there is a newly discovered area which has to be classified as the background or the moving object in a dynamic scene. 
The occlusion map $\Delta Z_{i}(\Tilde{u},\xi_{i-1}^{i})$ cannot be correctly calculated for $\Tilde{u}$, i.e. the pixels of the newly discovered area, because the warped pixel onto the previous image  $\text{w}(\Tilde{u},\xi_{i}^{i-1})$ is not in the image window $\Omega$. We define the newly discovered area as
\begin{equation} \label{def_nda}
    \Tilde{\Omega} = \{\Tilde{u} | \text{w}(\Tilde{u},\xi_{i}^{i-1})\not\in \Omega \}
\end{equation}
We suggest a method of predicting $A(\Tilde{u})$ {with the fast marching method  as the following:}  
\begin{equation} \label{def_nda_pred}
    A_i(\Tilde{u})  = \frac{\sum_{\delta u \not \in \Tilde{\Omega}} \left\{ A_i(\delta u) + \nabla_{\delta u} Z_i (\Tilde{u})\right\} }{\sum_{\delta u \not \in \Tilde{\Omega}}1}
\end{equation}
 The symbol $\delta u$ means the nearest pixels of $\Tilde{u}$ whose accumulated value $A(\delta u)$ has been calculated, and $\nabla_{\delta u}Z_i (\Tilde{u}) = Z_i (\Tilde{u}) - Z_i(\delta u)$ is the gradient of depth map with respect to $\delta u$. Because $A(u)$ has correspondence with the depth map, {we predict $A(\Tilde{u})$ with the gradient of the depth map and the occlusion accumulation map for the nearest pixel $\delta u$ adjoining the known area.} {In other words, we interpolate  $A(\Tilde{u})$ with the average of $A(\delta u) + \nabla_{\delta u} Z_i (\Tilde{u})$.} After that, we update the newly discovered area by $\Tilde{\Omega} \xleftarrow{} \Tilde{\Omega}-\{\delta u\}$, and repeat Eq. \eqref{def_nda_pred} until $\Tilde{\Omega} = \phi$. This process operates on two consecutive frames. Since the effect of the newly discovered area is not noticeable in two consecutive frames, we compared the masked image with some frame interval in Fig. \ref{fig:com_sfig4}.

\begin{figure}
 \centering
\begin{subfigure}{.22\textwidth}
  \centering
  \captionsetup{justification=centering}
  \includegraphics[width=.9\textwidth]{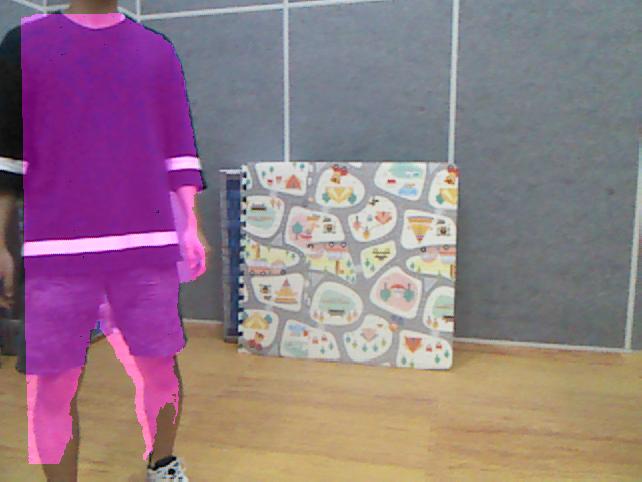}
  \caption{Masked image \newline (few frames ago)}
\end{subfigure} %
\begin{subfigure}{.22\textwidth}
  \centering
  \captionsetup{justification=centering}
  \includegraphics[width=.9\textwidth]{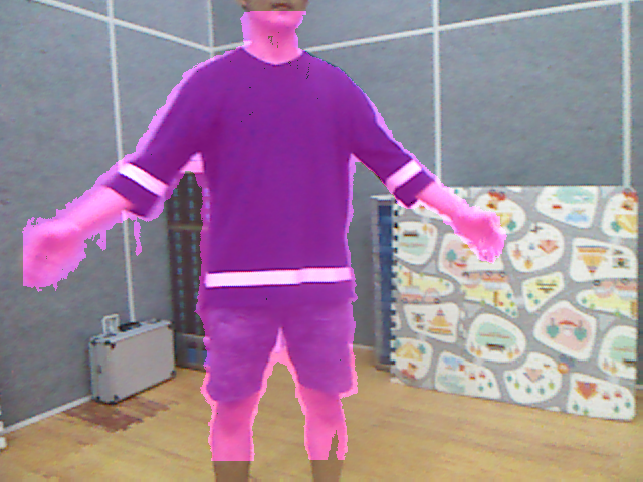}
  \caption{Masked image \newline (current frame)}
\end{subfigure} \\
\begin{subfigure}{.22\textwidth}
  \centering
  \includegraphics[width=.9\textwidth]{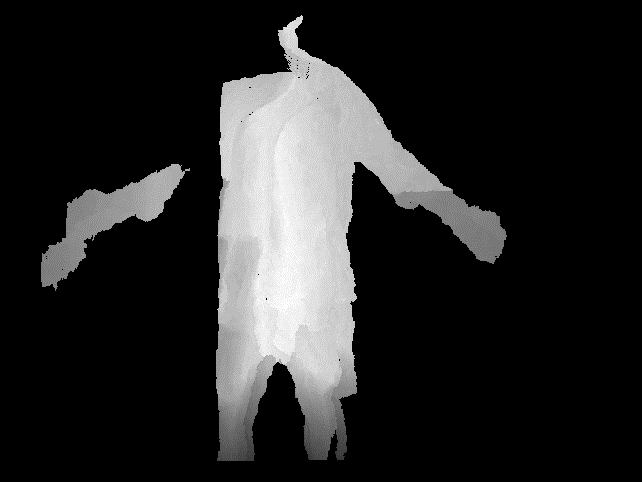}
  \caption{$A(u)$ without prediction}
\end{subfigure} %
\begin{subfigure}{.22\textwidth}
  \centering
  \includegraphics[width=.9\textwidth]{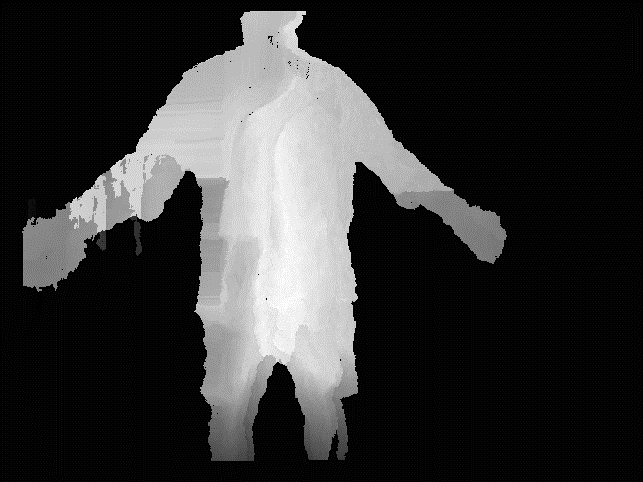}
  \caption{$A(u)$ with prediction}
\end{subfigure}

\caption{Example of occlusion prediction on the newly discovered area. When a new area is detected while the camera moves, there is no depth information in a few frames ago. Occlusion map has zero value on such area.  (c) shows the effect of the unmeasured depth values due to a newly discovered area. After executing  the occlusion prediction method, $A(u)$ has proper occlusion map values on the newly discovered area shown in (d) and the moving object detection result obtained from (d) is shown in (b).}
\label{fig:newly}
\vspace{-0.5cm}
\end{figure}

In the positive area of the predicted $A(u)$ which is considered as a moving object, the background depth has never been detected, because the background was continuously occluded by the moving object until the current image.
 The algorithm could fail to recognize such appearance of the background. Even if the background appears, $A(u)$ may not be lower than the threshold in Eq. \eqref{def_bg}. Then the truncation of Eq. \eqref{def_accum_re2} is triggered so the background recognition works properly. The result of the prediction is depicted in Fig. \ref{fig:newly}.
When there is no nearest moving object outside the predicted area, we remove $A(u)$ of the predicted area to prevent error propagation. Also, if there are small-sized moving object labels, we suppress them as they are usually negligible. 

\begin{figure*}[!ht]
\centering
\includegraphics[width=0.98\textwidth]{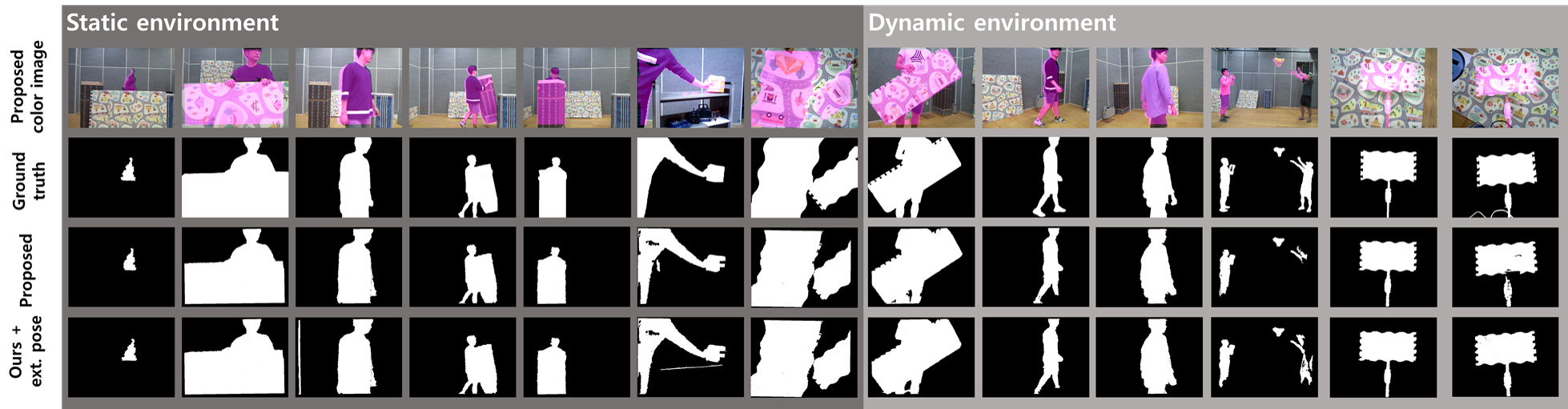}
\caption{Segmentation result for tested sequences in the order shown in Table.\ref{table:obj&VO}. The first row shows the masked image and the second row represents the ground truth. The result of the proposed method with VO and external pose estimation is presented orderly in the third and fourth rows {(except the TUM dataset for the reasons described in text)}. Since the depth is not measured on the boundary of the image, moving objects are not labeled in those areas.}
\label{fig:result}
\vspace{-0.5cm}
\end{figure*}

\section{Robust Pose Estimation} \label{pose_esti}

In this paper, we use DVO for camera pose measurement. Initially, we estimate the camera motion using only the sequence of RGB-D images. After detecting moving objects, the camera motion is estimated considering background mask $B(u)$.
The method using robust regression \cite{Kerl2013ROE} effectively estimate ego-motion over a small region of moving object, but it reaches a local minimum solution when the region of moving object is large as shown in Fig.\ref{fig:first}, because the background is confused with the moving object. On the other hand, the remaining moving object area after excluding the moving object detected up to the current frame is small enough to find upright camera pose. Thus, our method does not confuse the background and the moving object thanks to the accurate camera pose. 
 The camera pose estimation is achieved by minimizing the cost function as shown in Eq. \eqref{def_obj_fuc}. In the cost function, we use the bi-square norm shown in Eq. \eqref{def_bisquare} which completely ignores the effect of the outliers.
{ This property makes it possible to obtain the camera pose $\xi$ based on background pixels which have a relatively small residual.}

\begin{equation} \label{def_obj_fuc}
    \vspace{-0.25cm}
    \xi_i^{i+1} = \operatorname*{argmin}_{\xi}\sum_{u\in\Omega}  B_i (\text{w}(u,\xi)) 
    J_i(u,\xi)
    \vspace{-0.25cm}
\end{equation}
where
\begin{eqnarray} \label{def_residual}
    J_i(u,\xi) = \rho_{k_I}( \Delta I_{i}(u,\xi)) + \gamma \cdot \rho_{k_Z}( \Delta Z_{i}(u,\xi)) &&
\\ 
\label{def_bisquare}
    \rho_{k}(e) =
    \begin{cases}
        \frac{k^2}{6}\left\{ 1-\left[ 1 - (\frac{e}{k})^2 \right]^3 \right\}  & \text{for } |e| \leq k \\
        \frac{k^2}{6}              & \text{for } |e| > k
    \end{cases}&&
\end{eqnarray}

The symbol $I_i$ in Eq. \eqref{def_residual} is the $i$th color image and $k$  is the user-defined bi-square threshold. The residual values $\Delta I$ and $\Delta Z$ that are larger than $k$ do not influence the optimization process. In this paper, we use the levenberg-marquardt optimizer, and the parameter values $k_I = 48/255$, $k_Z = 0.5$, and $\gamma = 0.001$ are used.

\section{Experimental Results}

To evaluate our algorithm, we need RGB-D datasets in the presence of camera motion, ground truth data for moving objects, and camera trajectory. The dynamic object sequences of TUM dataset \cite{TUMdataset} contain the ground truth camera pose and RGB-D images, but the ground truth segmentation for moving object was not provided because the moving object such as a sitting person whose arm is moving could not be segmented clearly.
Thus, we used TUM dataset only to evaluate relative pose error. Additionally, we evaluate the performance of the proposed moving object detection and the pose estimation method with the dataset in \cite{leereal} and our dataset. 
The data sequences from \cite{leereal} are named with the prefix `ICSL' in \cref{table:obj&VO}.


We collected dataset with an RGB-D camera of Kinect V1, and obtained the ground truth pose of the camera using VICON. In addition, we manually obtained the ground truth segmentation of moving objects in pixel-wisely.

It contains 5 sequences for static camera, whose name starts with `$static$' and 4 sequences for the dynamic environment whose name starts with `$dynamic$' where camera motion and moving objects exist simultaneously. 
The sequences `$static\_tree$',`$static\_man$'  respectively contain  a tree and a man as a moving object. The sequence of `$static\_board$' contains a large-sized moving object which could be misrecognized as the background. In `$static\_destruct$', `$static\_construct$' sequences, there are situations where a building-shaped object is brought in and then removed. These sequences induce background model changes. In the sequence `$dynamic\_toss$', two people toss a doll to each other. It shows that multiple objects and fast objects can also be detected with our algorithm.
{The dataset and detailed information can be found here:}

\url{https://haram-kim.github.io/LARR-RGB-D-datasets/}

\begin{table*}[t]
\vspace{+0.3cm}
\caption{Object segmentation result (F1-score) \& relative pose error (RMSE)}
\vspace{-0.2cm}
\label{table:obj&VO}

\centering
\begin{tabular}{p{32mm} || C{10mm} | C{10mm} | C{10mm} | C{10mm} || C{8mm} | C{8mm} | C{8mm} | C{8mm} | C{8mm} | C{8mm}}
\toprule
\multirow{2}{*}{Sequences} & \multirow{2}{*}{\cite{MODTTVCMC}} & \multirow{2}{*}{\cite{MDTSA}}& \multirow{2}{*}{\begin{tabular}[c]{@{}c@{}}Proposed\\ method\end{tabular}}& \multirow{2}{*}{\begin{tabular}[c]{@{}c@{}}Ours with\\ ext. pose\end{tabular}}&
\multicolumn{2}{M{2cm}|}{Robust DVO}&
\multicolumn{2}{M{2cm}|}{Proposed method}& 
\multicolumn{2}{M{2cm}}{Joint VO-SF \cite{Jaimez2017FOSF}}\\ \cline{6-11}
& &
& &
&{tr.(m)} & {rot.($\tcdegree$)} 
&{tr.(m)} & {rot.($\tcdegree$)}
&{tr.(m)} & {rot.($\tcdegree$)}\\
\thickhline

static\_tree        & \textbf{0.9385}    & 0.8873     & 0.8821      &0.8892 
& 0.102    & 3.313   & 0.061    & \textbf{1.856}    & \textbf{0.032}    & 3.969\\

static\_board       & 0.9044    & \textbf{0.9256}     & 0.9182      &0.9177
&$\times$   &$\times$  & \textbf{0.292}    & \textbf{4.839}    & 0.584    & 10.916\\

static\_man         & 0.7350    & \textbf{0.8975}     & 0.8739      &0.8755
& 0.280    & 4.024   & 0.173    & 4.944    & \textbf{0.072}    & \textbf{1.045}\\

static\_destruct    & 0.5267    & 0.5565     & 0.8391      &\textbf{0.9038}& 0.535    & 4.818   & 0.336    & 4.495    & \textbf{0.131}    & \textbf{3.014}\\

static\_construct   & 0.3948    & 0.8362     & \textbf{0.8868}      &0.8396
& 0.809    & 10.433   & 0.153    & 3.821    & \textbf{0.078}    & \textbf{1.272}\\

ICSL\_place\_items    & 0.7139    & 0.6494     & 0.7204      & \textbf{0.7366}
& 0.041    & 1.082     & \textbf{0.030}      & \textbf{0.893}    & 0.094      & 5.379\\

ICSL\_two\_objects    & 0.5073    & 0.8256     & \textbf{0.8583}      & 0.8580
& 0.170    & 6.493     & \textbf{0.013}      & \textbf{0.344}    & - & - \\

\hline
dynamic\_board      & $\times$    & 0.6527     & \textbf{0.9264}    &{0.9247}
&$\times$   &$\times$  & \textbf{0.111}    & \textbf{1.939}   & 0.135  & 4.644\\

dynamic\_man1   & $\times$    & 0.4704     & {0.8123}    &\textbf{0.8287}
& 0.255    & 9.027   & \textbf{0.157}    & \textbf{4.108}   & 0.223    & 12.949\\

dynamic\_man2   & $\times$    & $\times$   & \textbf{0.8975}    &0.8955
& 0.646    & 11.286   & \textbf{0.166}    & \textbf{2.165}   & 0.206    & 6.180\\

dynamic\_toss   & $\times$      & 0.4395     & {0.7259}    &\textbf{0.7975}
& 0.635    & 8.546   & \textbf{0.324}    & \textbf{2.312}   & 0.378    & 4.028\\

ICSL\_fast\_object    & $\times$    & $\times$     & 0.8639      & \textbf{0.8779}
& 0.318    & 13.543     & \textbf{0.092}      & \textbf{3.603}    & -   & - \\

ICSL\_slow\_object    & $\times$    & 0.7424     & 0.8971      & \textbf{0.9142}
& 0.407    & 17.585     & \textbf{0.091}      & \textbf{3.779}    & -   & -\\

TUM\_sitting\_static    & & & &
& 0.037    & 0.972     & \textbf{0.035}      & \textbf{0.961}    & 0.045      & 1.699 \\

TUM\_sitting\_xyz     & & & &
& 0.078    & 2.027     & \textbf{0.073}      & \textbf{1.860}    & 0.205      & 4.066\\

TUM\_walking\_static    & & & &
& 0.370    & 2.581     & \textbf{0.217}      & \textbf{0.197}    & 0.249      & 4.173\\

TUM\_walking\_xyz       & & & &
& 0.974    & 15.871     & \textbf{0.259}      & \textbf{4.069}    & 0.659      & 12.370 \\

\bottomrule

\end{tabular}
\vspace{-0.6cm}
\end{table*}

The results of the object segmentation are shown in Table \ref{table:obj&VO} and Fig. \ref{fig:result}. We used the F1-score as a criterion for the moving object classification evaluation. The precision refers to how many of the pixels identified as moving objects are ground truth moving objects pixels and the recall refers to how many of the moving objects pixels in the image were correctly identified as moving objects. The F1-score is the harmonic mean of the precision and the recall. We calculated F1-score for each frame, and we averaged it in each sequence. The closer the score is to 1, the better the performance of moving object segmentation is.
We use the symbol `$\times$' when quantitative evaluation is meaningless due to extremely poor performance, {and  `-' for the algorithm that failed due to very few inliers in the corresponding sequence.}

We evaluated the segmentation performance of proposed method not only with robust DVO, but also with VICON data as an externally obtained camera pose. {The proposed method with camera pose of VICON data mostly performs better than the method with VO.}

In general static situations, all the compared algorithms perform well, but the performance of \cite{MODTTVCMC, MDTSA}  deteriorates in `$static\_destruct$' and `$static\_construct$' sequences, where the moving object appear at the beginning or the end of the video. The performance of \cite{MODTTVCMC} degrades even if the background and the moving object are different in color. The sequences of dynamic environments contain a difficult situation in that some newly discovered area found by the camera is covered by moving objects, thus the background of that area has never been revealed. The algorithms in \cite{MODTTVCMC, MDTSA} suffer in those situations so that the results were greatly affected.
Since we do not use the background model and predict $A(u)$ for the newly discovered area, our method distinguished moving objects stably in `$static\_destruct$', `$static\_construct$' and other dynamic environment sequences. {In the dataset of \cite{leereal}, our method correctly distinguishes the moving object from the background which have similar texture.}

\begin{figure}[!t]
\centering
\includegraphics[width=0.45\textwidth]{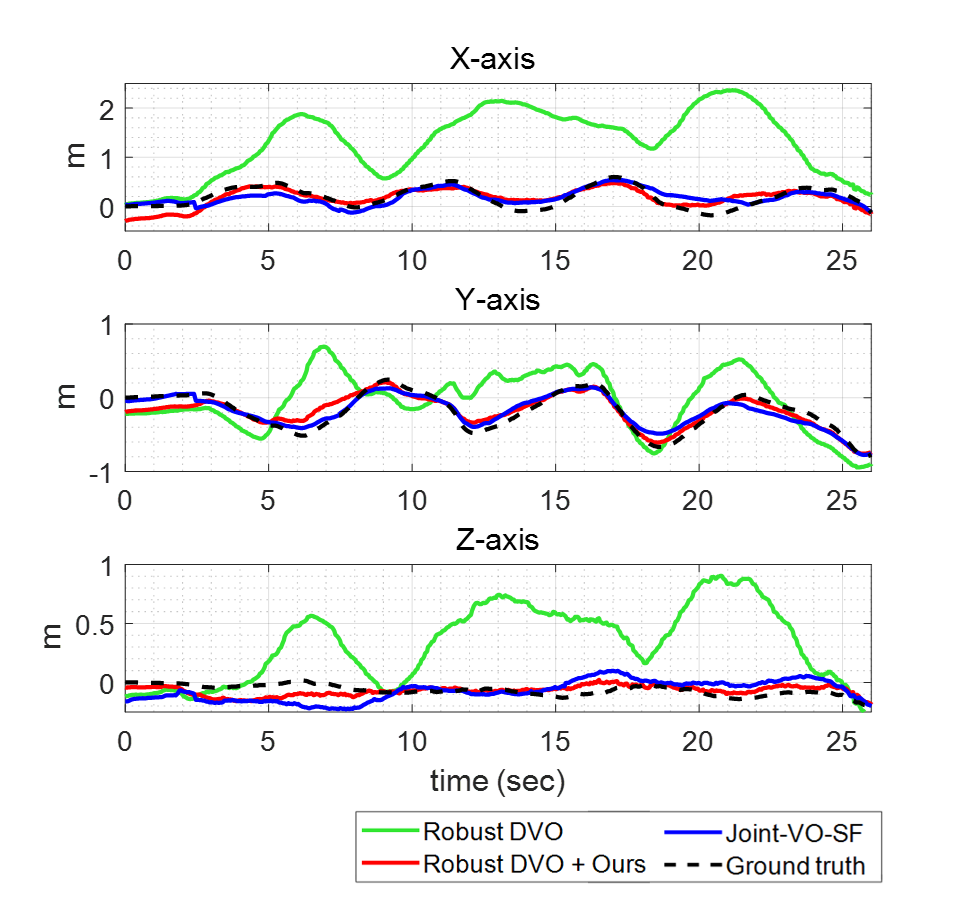}
\vspace{-0.1cm}
\caption{Translation estimation results on the sequence `$dynamic\_ man2$'. The robust DVO fails to estimate the ego-motion due to a moving object. When combined with our algorithm, the performance improves significantly.}
\label{fig:VO}
\vspace{-0.5cm}
\end{figure}

To evaluate the performance improvement of the VO when augmented with our algorithm, the translational and rotational relative pose error metric in \cite{TUMdataset} is adopted. We set the time parameter $\Delta =  150$, which means that we evaluate the drift per 5 seconds recorded at 30 Hz.
The results are shown in Table. \ref{table:obj&VO}, Fig. \ref{fig:VO}. 
Even though RMSE is calculated with the frames that do not have a moving object, our algorithm improves the estimation result of the robust DVO significantly {in various datasets, especially in dynamic environments.}

As a result of applying the joint VO-SF in \cite{Jaimez2017FOSF}, an algorithm based on DVO, it { mostly } performs better than DVO and shows smaller error than our proposed method in a \textit{ fixed} camera. For the $static\_board$ sequence where the moving object is dominant in the scene, the joint VO-SF method show large relative pose error than our method. {Interestingly, the results of $TUM\_sitting$ sequences show that the bi-square weight method is more robust than the Cauchy weight based method, which can be explained by their difference in outlier rejection intensity.} The proposed method, compared to robust DVO and Joint VO-SF in dynamic environment, performs better in both rotational and translational motion regardless of a moving object which dominates a scene.

As can be seen in Fig. \ref{fig:result}, although the moving objects are well separated, error exists due to the lack of texture in the background scene.
We expect that more accurate pose estimation in a dynamic environment can be achieved by applying other VO algorithms which show better performance on pose estimation in a static environment. 


\section{Conclusions}
We proposed a novel moving object detection method which utilizes occlusion accumulation and camera pose instead of a background model. To do this, we also presented the depth compensation on the unmeasured area and the occlusion prediction on the newly discovered area. Our method could detect the moving object which dominates the scene and improved the performance of robust DVO. In future work, we will combine the proposed method for robotic navigation tasks, which is designed to easily integrate with SLAM algorithms or obstacle avoidance algorithms.






\bibliographystyle{IEEEtran}
\bibliography{OA}

\end{document}